\documentclass[10pt,twocolumn,letterpaper]{article}

\usepackage[pagenumbers]{cvpr}

\usepackage{booktabs}

\usepackage{microtype}

\renewcommand{\paragraph}[1]{\vspace{.4em}\noindent\textbf{#1}}

\definecolor{cvprblue}{rgb}{0.21,0.49,0.74}
\usepackage[pagebackref,breaklinks,colorlinks,allcolors=cvprblue]{hyperref}

\def\paperID{397} 
\def\confName{3DV\xspace}
\def\confYear{2027\xspace}

\title{RLG-TPV: Radar- and LiDAR-Guided Tri-Perspective View Fusion \\ for Camera-Radar 3D Object Detection}

\author{Ahmet Mete Dokgoz \quad A. Enes Doruk \quad Hasan F. Ates\\
Department of Artificial Intelligence and Data Engineering, Ozyegin University, Istanbul, Turkiye\\
{\tt\small \{mete.dokgoz, enes.doruk\}@ozu.edu.tr \quad hasan.ates@ozyegin.edu.tr}
}

\begin{document}
\maketitle

\begin{abstract}
Tri-Perspective View (TPV) representations describe 3D scene structure through top, side, and front feature planes, but existing TPV lifting is primarily camera-based, leaving the depth of sampled image evidence ambiguous along projected camera rays. We propose RLG-TPV, a multimodal TPV framework for camera-radar 3D object detection in which radar and training-time LiDAR provide complementary geometric guidance during representation construction. A ray-guided deformable-attention lift weights sampled image features using LiDAR-supervised camera depth probabilities and radar frustum occupancy, while radar additionally refines the depth distribution before lifting. Because conventional radar provides limited elevation information, LiDAR-derived class-occupancy targets supervise the side and front planes during training; the corresponding heads are removed at inference, so deployment requires only cameras and radar. For temporal aggregation, Doppler-guided temporal fusion aligns past features using a motion field anchored by measured radar radial velocity, with gating that limits warping in regions without supported motion. An RCS-aware radar scatter further allows radar evidence to spread over spatial neighborhoods conditioned on radar cross section. On the nuScenes validation set, RLG-TPV achieves 0.4981 mAP and 0.5959 NDS, reducing orientation and velocity error by 31.9\% and 30.7\% relative to the published CRN baseline. Ablation studies show that ray-level geometric guidance is a major contributor to the final performance.
\end{abstract}

\section{Introduction}
\label{sec:intro}
\vspace{-0.2cm}
Camera-radar perception enables LiDAR-free inference for 3D object detection: cameras provide dense semantic information, while radar provides range, Doppler velocity, and complementary geometric measurements. Despite recent progress, camera-radar detection generally remains behind LiDAR-based approaches on standard autonomous-driving benchmarks~\cite{kitti,nuscenes,waymo,pointpillars,second,centerpoint}. Many camera-radar detectors construct a single Bird's-Eye View (BEV) representation~\cite{crn,rcbevdet,crtfusion}, which is convenient for detection but collapses the vertical dimension during feature construction. Tri-Perspective View (TPV) representations instead retain top ($xy$), side
($xz$), and front ($yz$) feature planes, preserving complementary spatial structure in which height remains explicit. This motivates the use of TPV as an intermediate representation for camera-radar 3D detection.

Retaining additional planes, however, does not by itself resolve how image features should be lifted into them. Existing TPV lifting is primarily camera-based and therefore inherits monocular depth ambiguity: a pixel is consistent with multiple 3D locations along its viewing ray. Once evidence
from these candidate locations has been aggregated into a spatial feature, later fusion can modulate that feature but cannot recover its original sample-level depth assignment. Radar is informative at this stage because its range and occupancy measurements can constrain the location of image
evidence along projected rays across all three TPV planes.  Radar, however, provides little direct elevation information and therefore cannot by itself constrain the vertical structure represented by the side and front planes. Training-time LiDAR provides a complementary source of geometry: it can supervise camera depth and the vertical structure of these planes without becoming an additional inference-time sensor.
These observations motivate a multimodal TPV formulation in which geometric cues are introduced while the corresponding spatial ambiguity is still explicit.

We propose RLG-TPV, a multimodal TPV framework for camera-radar 3D object detection. Its main component is a ray-guided deformable-attention lift~\cite{attention,deformabledetr}, in which LiDAR-supervised camera depth probabilities and radar frustum occupancy weight sampled image features along projected rays across all three TPV planes. Unlike Radar View Transform (RVT) formulations that incorporate radar occupancy within
Lift-Splat-Shoot~\cite{lss} frustum pooling, RLG-TPV introduces geometric guidance directly at the attention-sampling stage. Radar additionally refines the camera depth distribution before lifting, while LiDAR-derived class-occupancy targets supervise the side and front planes during training. The corresponding supervision is removed at inference, leaving a camera-radar detector. The radar branch further employs RCS-aware spatial scatter to distribute sparse radar evidence over local neighborhoods.

RLG-TPV also exploits radar Doppler during temporal aggregation. Combining past features without accounting for object motion can place observations of moving objects at inconsistent spatial locations. We therefore introduce
Doppler-Guided Temporal Fusion (DGTF), which uses measured radar radial velocity to initialize a motion field for aligning past fused features with the current frame. A learned residual corrects this field, while a motion-dependent gate limits warping in regions without supported motion. Radar therefore contributes both geometric guidance during spatial lifting and measured motion information during temporal fusion.

On the nuScenes validation set, RLG-TPV achieves 0.4981 mAP and 0.5959 NDS, with relative reductions of 31.9\% in orientation error and 30.7\% in velocity error compared with the published CRN baseline. Controlled lifting experiments indicate that the granularity at which geometric information enters the lift is important.

Our contributions are:
\begin{itemize}\itemsep0.15em

\item We formulate multimodal TPV lifting for camera-radar 3D detection, where radar range cues and training-time LiDAR geometry are incorporated according to the spatial information available from each sensor.

\item We introduce a ray-guided TPV lift in which LiDAR-supervised camera depth and radar frustum occupancy weight deformable-attention samples along projected camera rays across all three TPV planes. Radar additionally refines the depth distribution, while LiDAR-derived occupancy targets
provide training-time supervision for the vertical structure of the side and front planes without requiring LiDAR at inference.

\item We introduce Doppler-Guided Temporal Fusion (DGTF), which uses measured radar radial velocity to guide temporal alignment and applies motion-dependent gating to limit warping in regions without supported motion.

\item RLG-TPV reaches 0.4981 mAP and 0.5959 NDS on nuScenes val, while controlled ablations identify sample-level geometric guidance as the largest contributor among the investigated lifting variants.

\end{itemize}
\section{Related Work}
\label{sec:related}
\vspace{-0.2cm}

\textbf{Camera-radar view transformation and TPV lifting.}
Earlier camera-radar detectors such as CenterFusion~\cite{centerfusion} associate radar measurements with image-derived object hypotheses, while later approaches increasingly incorporate radar information into shared spatial representations or directly into the camera view transformation. CRN~\cite{crn} combines the camera depth distribution with radar occupancy during its Radar View Transform (RVT), and CRAB~\cite{crab} similarly uses radar cues to reduce depth ambiguity during backward projection. RaCFormer~\cite{racformer} incorporates radar-guided depth estimation within a query-based camera-radar detector. In parallel, BEVFormer~\cite{bevformer} replaces depth-based frustum pooling with deformable cross-attention from learnable spatial queries, while BEVDepth~\cite{bevdepth} retains an LSS-style transformation and uses LiDAR supervision to improve camera depth estimation. TPVFormer~\cite{tpvformer} extends attentive camera lifting to a tri-perspective representation composed of top, side, and front planes for camera-based 3D semantic occupancy prediction. These works provide complementary directions in radar-assisted depth reasoning and attentive multi-plane representation. RLG-TPV considers their combination for camera-radar 3D detection by incorporating radar occupancy and LiDAR-supervised depth directly into deformable-attention sampling across the TPV planes. LiDAR-derived occupancy is additionally used during training to supervise the side and front planes, where radar provides limited elevation information.

\textbf{Temporal fusion with radar motion cues.}
Temporal information has been used in camera perception to aggregate features across frames, including recurrent BEV fusion in BEVFormer~\cite{bevformer}, long-term temporal fusion in SOLOFusion~\cite{solofusion}, and object-centric temporal propagation in StreamPETR~\cite{streampetr}. Camera-radar methods also incorporate temporal information. CRN~\cite{crn} aggregates features from multiple temporal inputs, while CRT-Fusion~\cite{crtfusion} uses predicted velocity to align temporal BEV features. RaCFormer~\cite{racformer} models temporal dynamics through an implicit dynamic module that also exploits radar Doppler information. In RLG-TPV, measured radar Doppler is used explicitly as a motion cue for temporal feature alignment: past fused features are warped according to a radar-derived motion field, with a learned residual accounting for corrections not represented by the sparse measurements.

\textbf{Training-time LiDAR supervision and radar encoding.}
LiDAR can provide geometric information during training without being required as an inference-time sensor. BEVDepth~\cite{bevdepth} uses LiDAR to supervise camera depth estimation, while TPVFormer~\cite{tpvformer} is trained with sparse LiDAR semantic supervision for camera-only occupancy prediction. Knowledge-distillation approaches such as CRKD~\cite{crkd} and RCTDistill~\cite{rctdistill} further use LiDAR-derived information to supervise camera-radar models. Our use of LiDAR is limited to direct geometric supervision: it provides depth targets for the camera depth distribution and class-occupancy targets for the side and front TPV planes, while inference uses only cameras and radar.

\begin{figure*}[t]
\centering
\includegraphics[width=\textwidth]{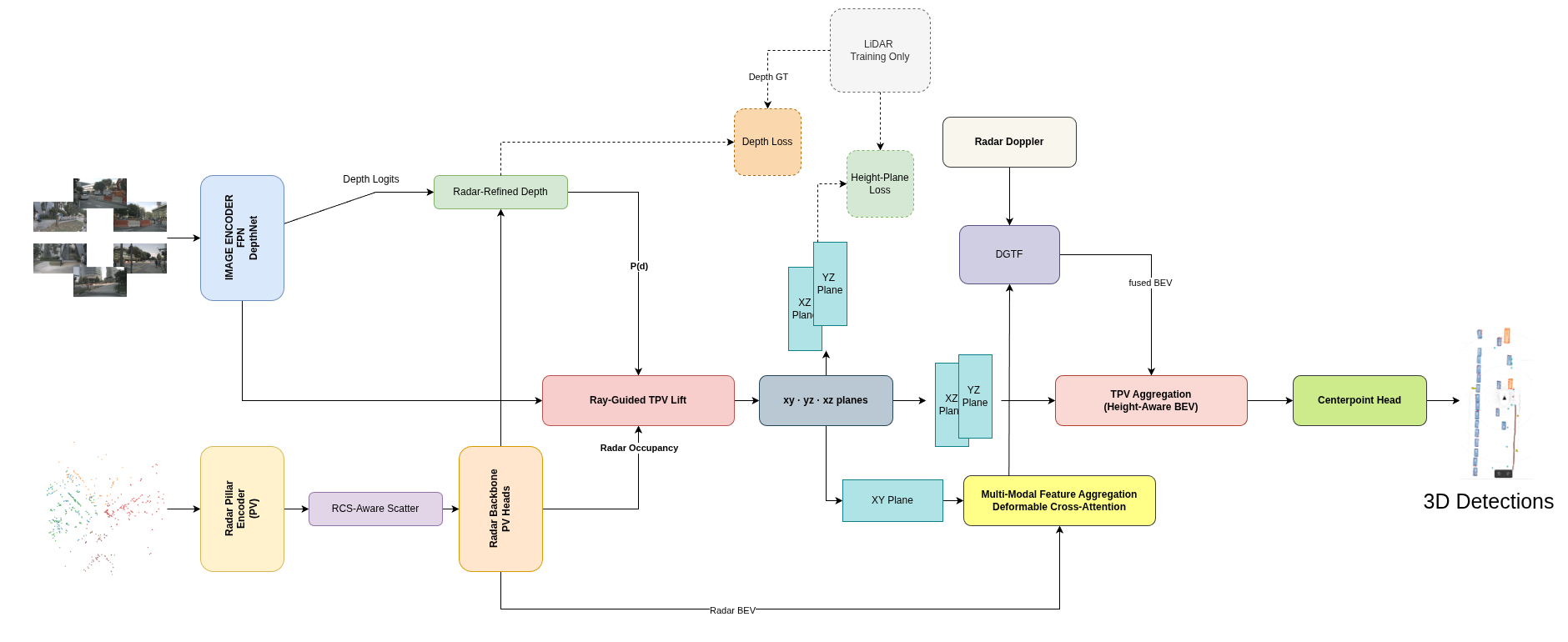}
\vspace{-0.6cm}
\caption{Overview of RLG-TPV. Camera features are lifted into TPV using depth- and radar-guided attention, fused with RCS-aware radar features, temporally aligned using Doppler, and aggregated into BEV for 3D detection. Dashed paths denote training-only LiDAR supervision.}
\label{fig:arch}
\end{figure*}

Radar encoding methods have also considered the spatial characteristics of sparse radar returns. RCBEVDet~\cite{rcbevdet} uses RCS as a prior when scattering radar features over multiple BEV locations rather than restricting each return to a single cell. RICCARDO~\cite{riccardo} models the spatial distribution of radar returns conditioned on object properties, providing another mechanism for representing radar measurements beyond isolated point locations. Our RCS-aware scatter follows the related motivation of allowing radar evidence to occupy spatial support rather than a single cell, but implements this as a residual multi-scale spreading operation controlled by RCS, applied to the scattered radar BEV grid before the radar backbone.

\vspace{-0.2cm}
\section{Method}
\label{sec:method}

\subsection{Framework Overview}
\label{sec:overview}
\vspace{-0.2cm}
RLG-TPV takes four temporally spaced camera-radar keyframes and predicts 3D bounding boxes for the current frame. Each keyframe contains six surround-view images and measurements from five radar sensors. As shown in \cref{fig:arch}, the pipeline contains a ResNet50~\cite{resnet}
camera backbone with DepthNet, a PointPillars~\cite{pointpillars} radar encoder, camera-radar fusion in BEV, and a CenterPoint head~\cite{centerpoint}. RLG-TPV modifies the representation construction between these components through multimodal TPV lifting, radar-aware encoding, and Doppler-guided temporal fusion.

For the current frame, camera features are lifted into three TPV planes: a ground plane $\mathbf{P}_{xy}$ and two height-preserving planes $\mathbf{P}_{xz}$ and $\mathbf{P}_{yz}$. Instead of frustum pooling, RLG-TPV samples image features along projected camera rays using deformable attention. Each sample is weighted by the camera depth probability $P(d\mid u,v)$
and radar frustum occupancy before aggregation, so geometric evidence influences the lift while the depth ambiguity is still explicit (\cref{sec:lift}). Radar additionally refines the camera depth logits before lifting, making the resulting depth distribution conditional on both modalities.

The radar branch represents sparse returns using an RCS-aware scatter that distributes radar features over local spatial neighborhoods conditioned on radar cross section (\cref{sec:radar}). For each keyframe, the camera
$\mathbf{P}_{xy}$ plane is fused with the radar BEV using the multi-modal feature aggregation (MFA) module of CRN~\cite{crn}. The fused $xy$ features from past keyframes are then aligned to the current frame using Doppler-guided temporal fusion, whose motion field is anchored by radar-derived velocity measurements.

The current-frame side and front planes retain vertical structure that is not directly measured by conventional radar. We therefore supervise these planes during training with class-labeled LiDAR occupancy (\cref{sec:lidar}), while LiDAR depth also supervises the camera depth distribution. LiDAR is used only for training and is absent at inference.

Finally, the current TPV planes and the temporally fused $xy$ representation are combined into a common height-resolved representation and collapsed along the vertical dimension to form the BEV feature consumed by the CenterPoint head. The detector then predicts the class, 3D location, size, orientation, and velocity of each object in the current frame.

\textbf{Training objective.} The complete objective is
\begin{equation}
    \mathcal{L}
    =
    \mathcal{L}_{\mathrm{det}}
    + 3.0\,\mathcal{L}_{\mathrm{depth}}
    + 0.5\,\mathcal{L}_{\mathrm{plane}},
\end{equation}
where $\mathcal{L}_{\mathrm{det}}$ is the CenterPoint detection loss,
$\mathcal{L}_{\mathrm{depth}}$ supervises the refined camera depth
distribution and $\mathcal{L}_{\mathrm{plane}}$ supervises the side and front TPV planes. The loss weights compensate for scale differences across the detection, depth, and plane-supervision objectives.

\subsection{Ray-Guided Multimodal TPV Lift}
\label{sec:lift}
\vspace{-0.2cm}

\begin{figure}[t]
\centering
\includegraphics[width=\columnwidth]{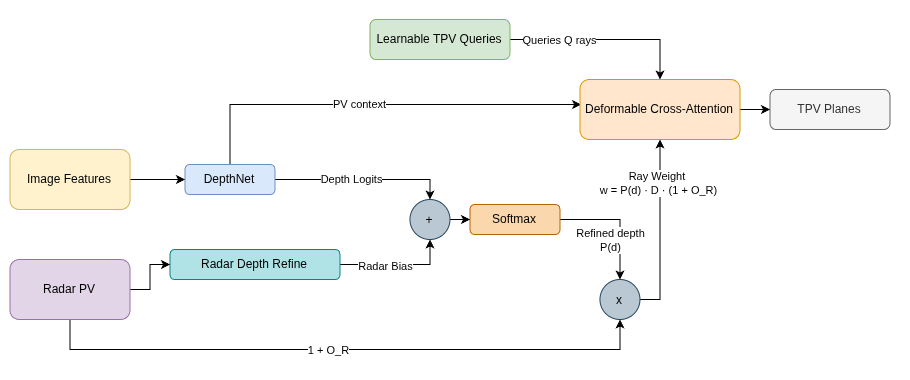}
\vspace{-0.5cm}
\caption{Radar-refined depth and ray-guided lifting. Radar features bias the camera depth logits via $b_R$ and provide occupancy $O_R$ for per-ray sample weighting before deformable-attention aggregation into the three TPV planes; $D$ in the figure denotes $D_{\mathrm{bins}}$.}
\label{fig:raylift}
\end{figure}

A pixel is consistent with multiple 3D locations along its viewing ray, making camera-only lifting depth-ambiguous. RLG-TPV introduces geometric guidance before aggregation: the LiDAR-supervised camera depth distribution and radar frustum occupancy weight individual image samples along projected
rays, while deformable attention determines the sampled feature content. The lift produces the three TPV planes $\mathbf{P}_{xy}$, $\mathbf{P}_{xz}$, and $\mathbf{P}_{yz}$, as summarized in \cref{fig:raylift}.

Each TPV cell fixes two spatial coordinates while leaving the third unspecified. For example, an $xy$ cell fixes $(x,y)$ but not $z$, whereas a $yz$ cell fixes $(y,z)$ but not $x$. We therefore place \(K\) reference points along the missing axis of each cell, yielding concrete 3D locations that \(K\) may differ across the TPV planes according to the spatial extent of the corresponding missing axis. These points are projected into the surrounding cameras and used to sample
image features by deformable cross-attention~\cite{deformabledetr}. The camera-frame depth of each projected point is also used to query the depth distribution and radar frustum occupancy.

\textbf{Radar-refined depth.}
Because the ray-guided lift depends on the camera depth distribution, radar range information is used to refine this distribution before lifting. Let \(\boldsymbol{\ell}\) denote the DepthNet logit volume, with \(\ell(d,u,v)\) its entry at depth bin \(d\) and image location \((u,v)\), and let \((C_R,O_R)\) denote the radar perspective-view context and occupancy. A lightweight module predicts a radar-conditioned bias $b_R(d,u)$ from the radar context $C_R$ that is added before the depth softmax:
\begin{equation}
  P(d\mid u,v)
  =
  \mathrm{softmax}_d\bigl(\ell(d,u,v)+b_R(d,u)\bigr).
  \label{eq:rrd}
\end{equation}
Since conventional radar provides little direct elevation information, $b_R(d,u)$ is shared across image rows. The refinement is trained through the existing LiDAR depth supervision and introduces no additional loss term.

\textbf{Per-ray depth and radar guidance.}
Standard deformable attention learns sampling offsets and attention weights from image features but does not explicitly encode range support along the camera ray. We therefore assign an additional geometric weight to each projected sample. For a reference point projected into camera $i$ at image position $(u,v)$ and camera-frame depth $d$,
\begin{equation}
  w(i,u,v,d)
  =
  P(d\mid u,v)\,
  D_{\mathrm{bins}}\,
  \bigl(1+O_R(d,u)\bigr),
  \label{eq:weight}
\end{equation}
where all image-plane quantities are evaluated in camera \(i\): \(P(d\mid u,v)\) is the radar-refined, LiDAR-supervised depth distribution, \(D_{\mathrm{bins}}\) is the number of depth bins, and \(O_R(d,u)\) denotes radar frustum occupancy obtained by projecting radar returns into the camera view and discretizing their ranges using the same depth bins as the camera depth distribution. Since conventional radar provides limited elevation information, \(O_R(d,u)\) is represented in depth—column space and shared across image rows. Writing $(u_{ik},v_{ik},d_{ik})$ for the image position and camera-frame depth of reference point $k$ projected into camera $i$, we abbreviate the weight of that sample as $w_{ik}\triangleq w(i,u_{ik},v_{ik},d_{ik})$.

The same weighting is applied to all three TPV planes. Radar therefore provides range guidance to the side and front planes as well, although it does not directly measure their elevation dimension. The additive-one term preserves camera evidence where no radar return is available, while $D_{\mathrm{bins}}$ normalizes the depth term around one for a uniform depth distribution.

The feature associated with TPV cell $c$ is then written as
\begin{equation}
  \mathbf{t}[c]
  =
  \frac{1}{N_{\mathrm{valid}}}
  \sum_{i=1}^{6}\sum_{k=1}^{K}
  w_{ik}
  \sum_{p=1}^{N_{\mathrm{pts}}}
  \alpha_{p}\,
  \mathbf{V}\!\left(\mathbf{g}_{ik}+\Delta_{p}\right),
  \label{eq:attend}
\end{equation}

where $\mathbf{g}_{ik}=(u_{ik},v_{ik})$ is the projected image position of the reference point $k$ in camera $i$, $\mathbf{V}$ is the multi-scale image value map from which features are bilinearly sampled, and $p=1,\dots,N_{\mathrm{pts}}$ indexes the deformable sampling points, whose learned offsets and attention weights are $\Delta_p$ and $\alpha_p$ with $\sum_{p=1}^{N_{\mathrm{pts}}}\alpha_p = 1$. The sum runs only over reference points that fall inside the image plane of camera $i$, and $N_{\mathrm{valid}}$ is the number of such valid camera-reference-point pairs so that cells observed by several cameras are not over-counted. The geometric weight $w_{ik}$ scales the sampled contribution outside the attention softmax, allowing range evidence to modulate the feature without replacing the learned
attention distribution.

\textbf{Multi-keyframe lift.}
The $xy$ plane is constructed for the current and past keyframes and fused with the corresponding radar BEV using the MFA module of CRN~\cite{crn}. Past fused features are subsequently aligned to the current frame by the Doppler-guided temporal fusion described in \cref{sec:radar}. The
$\mathbf{P}_{xz}$ and $\mathbf{P}_{yz}$ planes are constructed only for the current keyframe, since they enter the final TPV aggregation directly.

\textbf{TPV aggregation.}
After temporal fusion, let $\tilde{\mathbf{P}}_{xy}$ denote the temporally fused camera-radar $xy$ representation. The three planes are then broadcast to a common volume and combined as
\begin{equation}
  \mathbf{F}(x,y,z)
  =
  \tilde{\mathbf{P}}_{xy}[x,y]
  +
  \mathbf{P}_{xz}[x,z]
  +
  \mathbf{P}_{yz}[y,z],
\end{equation}
where each plane is indexed by its own two coordinates in the order given by its name. The height dimension is then collapsed by a learned depthwise 3D operation to form the BEV representation consumed by the detection head.

\subsection{Radar Encoding and Doppler-Guided Temporal Fusion}
\label{sec:radar}
\vspace{-0.1cm}

Radar provides two complementary signals to RLG-TPV: sparse spatial measurements and radial motion. We modify the radar representation with an RCS-conditioned spatial scatter and use measured Doppler as an anchor for temporal feature alignment.

\textbf{RCS-aware scatter.}
Standard pillar encoding assigns each radar pillar feature to a single cell of the BEV grid, which preserves the sparse structure of the measurements but provides limited spatial support around each return. We therefore add a residual multi-scale scatter whose spatial extent is conditioned on radar cross section (RCS). The scatter acts on the BEV pillar grid, immediately after pillar scattering and before the radar backbone. Let $\mathbf{X}$ denote the scattered radar BEV feature map and $\mathbf{R}$ the corresponding per-cell RCS map. The refined feature is
\begin{equation}
\begin{aligned}
  \mathbf{X}_{\mathrm{out}}
  &= \mathbf{X}
  + \mathrm{proj}\!\Bigl(
  \textstyle\sum_{m=1}^{S}
  \mathrm{Blur}_m\bigl(\mathbf{X}\odot\boldsymbol{\gamma}_m\bigr)
  \Bigr), \\
  \boldsymbol{\gamma}
  &= \mathrm{softmax}_S\!\bigl(h_R(\mathbf{R})\bigr),
\end{aligned}
\end{equation}
where $\mathrm{Blur}_m$, $m=1,\dots,S$, denotes fixed Gaussian kernels at different spatial scales and $\mathrm{proj}$ is a $1\times1$ projection. The gate $\boldsymbol{\gamma}\in\mathbb{R}^{S\times H\times W}$ is produced by a small convolutional predictor $h_R$ applied to the RCS map and normalized over the $S$ scales at every cell; $\boldsymbol{\gamma}_m$ is its $m$-th scale map, and $\odot$ broadcasts this single-channel map over the feature channels of $\mathbf{X}$. The residual formulation retains the original sparse radar feature while allowing RCS to modulate how much additional support is distributed to neighboring cells.

\textbf{Doppler-guided temporal fusion.}
Temporal aggregation can misalign features from moving objects when past BEV representations are combined without object-motion compensation. Radar provides sparse measured velocity cues that can anchor the motion of temporal BEV features. Let $\mathbf{f}_j$ denote the fused BEV feature of past keyframe
$j$ and $\mathbf{v}_j$ its BEV motion map constructed from the compensated radar velocity components. We define the alignment field as
\begin{equation}
  \boldsymbol{\phi}_j
  =
  \underbrace{s_j\,
  \mathrm{Smooth}(\mathbf{v}_j)}_{\text{Doppler anchor}}
  +
  \underbrace{\psi\bigl([\mathbf{f}_j,\mathbf{v}_j]\bigr)}
  _{\text{learned residual}},
  \label{eq:flow}
\end{equation}
where $\mathrm{Smooth}$ is a fixed Gaussian smoothing that extends the sparse radar measurements over their spatial neighborhood,
$s_j$ is a learnable scale, and $\psi$ predicts a residual correction. The Doppler term therefore provides an initial motion cue, while the learned residual accounts for motion that is not fully described by the sparse radial measurements.

To avoid applying motion compensation where radar does not indicate meaningful motion, the field is modulated by a Doppler-dependent gate:
\begin{equation}
  \boldsymbol{\phi}_j
  \leftarrow
  \boldsymbol{\phi}_j\,
  \sigma\!\left(
  \frac{\lVert\mathrm{Smooth}(\mathbf{v}_j)\rVert-\tau_v}{T}
  \right),
\end{equation}
where $\sigma$ is the logistic sigmoid applied per cell, $\tau_v$ is the speed threshold below which motion is treated as unsupported, and $T$ is a temperature that controls how sharply the gate transitions around $\tau_v$. The resulting field warps each past fused BEV toward the current frame. An occupancy gate further suppresses contributions from regions with weak feature support, after which the aligned past features and the current feature are combined by the temporal reduction module.

\subsection{Cross-Modal Height-Plane Supervision}
\label{sec:lidar}
\vspace{-0.2cm}

Radar provides range, azimuth, and Doppler measurements but little direct elevation information. Although radar range guidance is used during lifting across all three TPV planes, it does not directly supervise the vertical structure represented by the $\mathbf{P}_{xz}$ and $\mathbf{P}_{yz}$ planes. We therefore use LiDAR during training as an additional source of 3D geometry. LiDAR-derived
class-occupancy targets supervise the two height planes before TPV aggregation, while inference remains camera-radar only.

For each training keyframe, LiDAR points are transformed to the ego frame and assigned the class of the annotated 3D box containing them; points outside annotated objects are ignored. The labeled points are projected onto the $xz$ and $yz$ planes to construct class-occupancy targets $G_{xz}$ and $G_{yz}$. Cells without a labeled LiDAR return are excluded
from the loss. A lightweight classifier attached to each height plane is trained with
\begin{equation}
  \mathcal{L}_{\mathrm{plane}}
  =
  \mathrm{CE}(\hat{Y}_{xz},G_{xz})
  +
  \mathrm{CE}(\hat{Y}_{yz},G_{yz}),
\end{equation}
where $\hat{Y}_{xz}$ and $\hat{Y}_{yz}$ denote the corresponding class predictions. The auxiliary classifiers are removed after training, so this supervision introduces no additional inference-time sensor or prediction head.

The supervision is applied directly to the side and front TPV features before their aggregation into BEV, where the vertical coordinate is still explicit. This complements the LiDAR-supervised camera depth distribution: depth supervision constrains the placement of image evidence along camera
rays, while the plane targets provide an additional training signal for the vertical structure represented by the height planes.

\section{Experiments}
\label{sec:exps}
\vspace{-0.1cm}
\subsection{Experimental Setup}
\label{sec:setup}
\vspace{-0.1cm}

\begin{table*}[t]
\centering
\caption{Comparison with camera (C) and camera-radar (C+R) methods on nuScenes val at $256\times704$. Published CRN is shown here; its released implementation is used for ablations. Higher is better for mAP/NDS and lower for the error metrics.}
\vspace{-0.2cm}
\label{tab:sota}
\scriptsize
\setlength{\tabcolsep}{3.5pt}
\begin{tabular}{lccccccccc}
\toprule
Method & Input & Backbone & mAP & NDS & mATE & mASE & mAOE & mAVE & mAAE \\
\midrule
CenterFusion~\cite{centerfusion}
& C+R & DLA-34
& 0.332 & 0.453 & 0.649 & 0.263 & 0.535 & 0.540 & \textbf{0.142} \\

BEVDepth~\cite{bevdepth}
& C & ResNet-50
& 0.351 & 0.475 & 0.639 & 0.267 & 0.479 & 0.428 & 0.198 \\

CRAFT~\cite{craft}
& C+R & DLA-34
& 0.411 & 0.517 & 0.494 & 0.276 & 0.454 & 0.486 & 0.176 \\

CRN~\cite{crn}
& C+R & ResNet-50
& 0.490 & 0.560 & 0.487 & 0.277 & 0.542 & 0.344 & 0.197 \\

RCBEVDet~\cite{rcbevdet}
& C+R & ResNet-50
& 0.453 & 0.568 & 0.486 & 0.285 & 0.404 & 0.220 & 0.192 \\

CRT-Fusion~\cite{crtfusion}
& C+R & ResNet-50
& 0.500 & 0.572 & 0.499 & 0.277 & 0.531 & 0.261 & 0.192 \\

HyDRa~\cite{hydra}
& C+R & ResNet-50
& 0.494 & 0.585 & \textbf{0.463} & 0.268 & 0.478 & 0.227 & 0.182 \\

\textbf{RLG-TPV (ours)}
& C+R & ResNet-50
& 0.498 & 0.596 & 0.470 & 0.275 & \textbf{0.369} & 0.238 & 0.179 \\

RaCFormer~\cite{racformer}
& C+R & ResNet-50
& \textbf{0.541} & \textbf{0.613}
& 0.478 & \textbf{0.261} & 0.449 & \textbf{0.208} & 0.180 \\
\bottomrule
\end{tabular}
\end{table*}

\textbf{Dataset and evaluation metrics.}
All experiments use nuScenes~\cite{nuscenes}, with 700 training scenes (28,130 samples), 150 validation scenes (6,019 samples), and 10 detection classes. Evaluation follows the standard nuScenes protocol, reporting mAP over BEV center-distance thresholds $\{0.5,1.0,2.0,4.0\}$~m and the nuScenes Detection Score (NDS). We additionally report mean Average Translation Error (mATE), mean Average Scale Error (mASE), mean Average
Orientation Error (mAOE), mean Average Velocity Error (mAVE), and mean Average Attribute Error (mAAE). Higher values are better for mAP and NDS, whereas lower values are better for the five error metrics.

\textbf{Implementation.}
Images are resized to $256\times704$. We use a multi-keyframe temporal window at one-second spacing. Models are trained for 24 epochs using AdamW~\cite{adamw} with learning rate $2\times10^{-4}$, weight decay $10^{-4}$, cosine decay with one warm-up epoch, batch size 4, fp16 training, and gradient checkpointing in the TPV lift. The image backbone is an ImageNet-pretrained ResNet-50~\cite{resnet}. The TPV queries and the attentive lift are randomly initialized, whereas the radar-refined depth and the RCS-aware scatter are zero-initialized so that the added paths start from the baseline behavior, and the DGTF residual starts at a negligible scale so that the temporal fusion starts at the measured-Doppler warp. Controlled comparisons use the released CRN~\cite{crn} implementation with the same keyframe setting; its released configuration reaches 0.4725 mAP / 0.5617 NDS, below the published mAP of 0.490 while slightly exceeding the published NDS of 0.560, so we use the released implementation for internal ablations and the published value only for the comparison with prior methods. Incremental configurations in \cref{tab:tpv} are warm-started from the preceding row and retrained for the complete 24-epoch schedule. Evaluation uses EMA weights and a single deterministic forward pass without test-time augmentation, class-balanced resampling, or ensembling.

\subsection{Comparison with Prior Methods}
\label{sec:sota}
\vspace{-0.2cm}

\cref{tab:sota} compares RLG-TPV with published camera and camera-radar detectors on nuScenes val, using the published configuration of each competing method. Its 0.596 NDS is the second highest in the table and the highest among all models trained from ImageNet initialization with a 24-epoch schedule, exceeding HyDRa by 0.011 and CRT-Fusion by 0.024, while its 0.498 mAP is above the published CRN result (0.490) and HyDRa (0.494). The lead is clearer in NDS than in mAP because half of NDS reflects how accurate the predicted boxes are rather than how many objects are found.

Orientation is the most distinct result: 0.369 mAOE is the lowest value reported, $8.7\%$ below the next entry (RCBEVDet, 0.404) and $31.9\%$ below the published CRN baseline. Velocity error moves in the same direction, 0.238 against 0.344 for CRN, and translation error is the second lowest at 0.470. These improvements are consistent with the height-aware and motion-aware design of RLG-TPV: the side and front planes retain additional vertical structure, while DGTF incorporates radar-derived motion cues during temporal alignment. We do not attribute an individual metric to an individual module from this table alone; the controlled experiments below separate the mechanisms.

RLG-TPV does not reach the mAP of RaCFormer, which uses a query-based decoder, nuImages pretraining and 36 epochs, although it achieves a lower orientation error ($0.369$ against $0.449$). Comparing per-class AP against the released CRN implementation, RLG-TPV improves nine of the ten nuScenes classes, led by pedestrian ($+7.8$ AP), with car, bus and truck also improving consistently and motorcycle the only regression ($-1.3$ AP), so the gains span small unreflective objects and large radar-reflective ones alike.

\subsection{Analysis of Ray-Guided TPV Lifting}
\label{sec:lift_abl}
\vspace{-0.2cm}

\cref{tab:lift} examines where geometric information has to enter an attentive lift. Without a depth prior, samples taken at different depths along a projected camera ray carry no indication of which depths the scene actually supports; once they are aggregated into a spatial cell, a radar gate can rescale the resulting feature but can no longer tell which sampled depth produced it. This is why per-cell radar modulation recovers little over unguided attention, whereas applying the depth distribution and the radar occupancy to the individual samples recovers approximately $87\%$ of the difference between unguided attentive lifting and LSS+RVT. These rows also differ in their head---the unguided and per-cell variants use the three-plane configuration with a 3D-convolutional head---so this figure bounds rather than isolates the effect of guidance granularity. The isolated estimate is the ray-guidance removal in \cref{tab:abl}.

\begin{table}[hbpt]
\caption{Camera-lift ablation on nuScenes val. Radar and MFA are fixed. Unguided variants use three TPV planes with a 3D-conv head, while per-ray variants use xy-only CenterPoint.}
\vspace{-0.2cm}
\centering
\label{tab:lift}
\scriptsize
\setlength{\tabcolsep}{4pt}
\begin{tabular}{lcc}
\toprule
Lift variant & mAP & NDS \\
\midrule
LSS + RVT (warm-start)
& 0.4886 & 0.5842 \\
\midrule
Unguided attentive
& 0.4266 & 0.5321 \\
Per-cell radar gate
& 0.4311 & 0.5354 \\
{\bfseries\boldmath Per-ray $P(d)\times$radar, xy only}
& \textbf{0.4804} & \textbf{0.5782} \\
XYONLY (keyframe-only)
& 0.4407 & 0.5422 \\
\bottomrule
\end{tabular}
\end{table}

The keyframe-only configuration degrades substantially relative to the multi-frame per-ray variant, so temporal image evidence is likewise an important part of the lifting configuration. As a further control, neither widening the fused BEV nor attaching height-aware supervision after BEV fusion produced a resolved improvement over the single-plane baseline. Together, these results motivate applying geometric constraints while the relevant spatial ambiguity is still represented, rather than after projection and aggregation.

\subsection{Incremental Construction of RLG-TPV}
\label{sec:tpv_incremental}
\vspace{-0.2cm}

The progression in \cref{tab:tpv} separates the ray-guided representation change from the refinements that follow. We list the single-plane warm start alongside the released CRN baseline: the xy-only ray-guided lift falls $0.0082$ mAP below it, and the final model gains $+0.0095$ mAP / $+0.0117$ NDS over it, against $+0.0256$ / $+0.0342$ over CRN. The largest increase after ray guidance comes from the move to the complete multi-plane configuration, but that row introduces the side and front planes, their supervision and the RCS-aware radar representation at once, while also removing the auxiliary velocity module used in the single-plane warm start. We therefore read it as evidence for the combined multi-plane configuration rather than for any single component.

\begin{table}[hbpt]
\centering
\caption{Incremental construction of RLG-TPV on nuScenes val. From the ray-guided lift onward, each row is warm-started from the previous configuration and retrained for 24 epochs; changes within $\pm0.005$ run-to-run variation are discussed in the text. The warm start is the LSS\,+\,RVT row of \cref{tab:lift}.}
\vspace{-0.2cm}
\label{tab:tpv}
\scriptsize
\setlength{\tabcolsep}{3pt}
\begin{tabular}{lcc}
\toprule
Configuration & mAP & NDS \\
\midrule
CRN baseline (released code)
& 0.4725 & 0.5617 \\

Single-plane warm start (LSS\,+\,RVT)
& 0.4886 & 0.5842 \\
\midrule

Ray-guided TPV lift, xy only
& 0.4804 & 0.5782 \\
\midrule

+ height planes, plane sup., RCS scatter
& 0.4920 & 0.5882 \\

\quad target: GT boxes $\rightarrow$ LiDAR occ.
& 0.4925 & 0.5872 \\

+ radar-refined depth (RRD)
& 0.4932 & 0.5893 \\

+ sampling density ($K$: $4\rightarrow8/16$)
& 0.4968 & 0.5919 \\

\textbf{+ Doppler-guided temporal fusion (final)}
& \textbf{0.4981} & \textbf{0.5959} \\
\bottomrule
\end{tabular}
\end{table}
\vspace{-0.2cm}

Changing the height-plane target from box rasterization to class-labeled LiDAR occupancy produces no resolved difference, which separates the \emph{source} of the target from its \emph{placement}: a measured target and one derived from annotations behave alike when attached to the same planes. This suggests that the benefit primarily arises from supervising the height planes, while the exact source of the sparse target is less critical in the current configuration. These rows carry one caveat: the offline targets follow horizontal flipping but not the rotation and scaling of the BEV augmentation, which misaligns target and feature on augmented samples and plausibly explains why the term shows no isolated gain. We use them to support the placement of the geometric constraint, not to claim a standalone gain from it.

RRD and DGTF change mAP within the run-to-run variation and are not treated as resolved standalone gains, although DGTF changes NDS more than mAP, as expected from a module that improves box quality rather than the number of detections. Increasing the sampling density also stays within that variation here, although removing it from the final model does not (\cref{tab:abl}): with only four reference points, consecutive height-plane samples are about $34$\,m apart along their missing spatial axis, making the attentive approximation particularly sparse.

\subsection{Component Ablations}
\label{sec:abl}
\vspace{-0.2cm}

The removal study in \cref{tab:abl} supports the same hierarchy as the incremental
construction. Removing temporal context or ray-level guidance produces the largest architecture-specific degradations, indicating that temporal evidence and depth-aware sampling are structural parts of the final representation rather than minor refinements. Reducing the sampling density has a smaller but still resolved effect. RCS-aware scatter changes NDS more
strongly than mAP, suggesting that its spatial treatment of radar evidence has a larger effect on prediction quality than on the number of detections. In contrast, removing RRD has little effect on mAP, consistent with its small incremental change.

\begin{figure*}[t]
\centering
\includegraphics[width=0.8\textwidth]{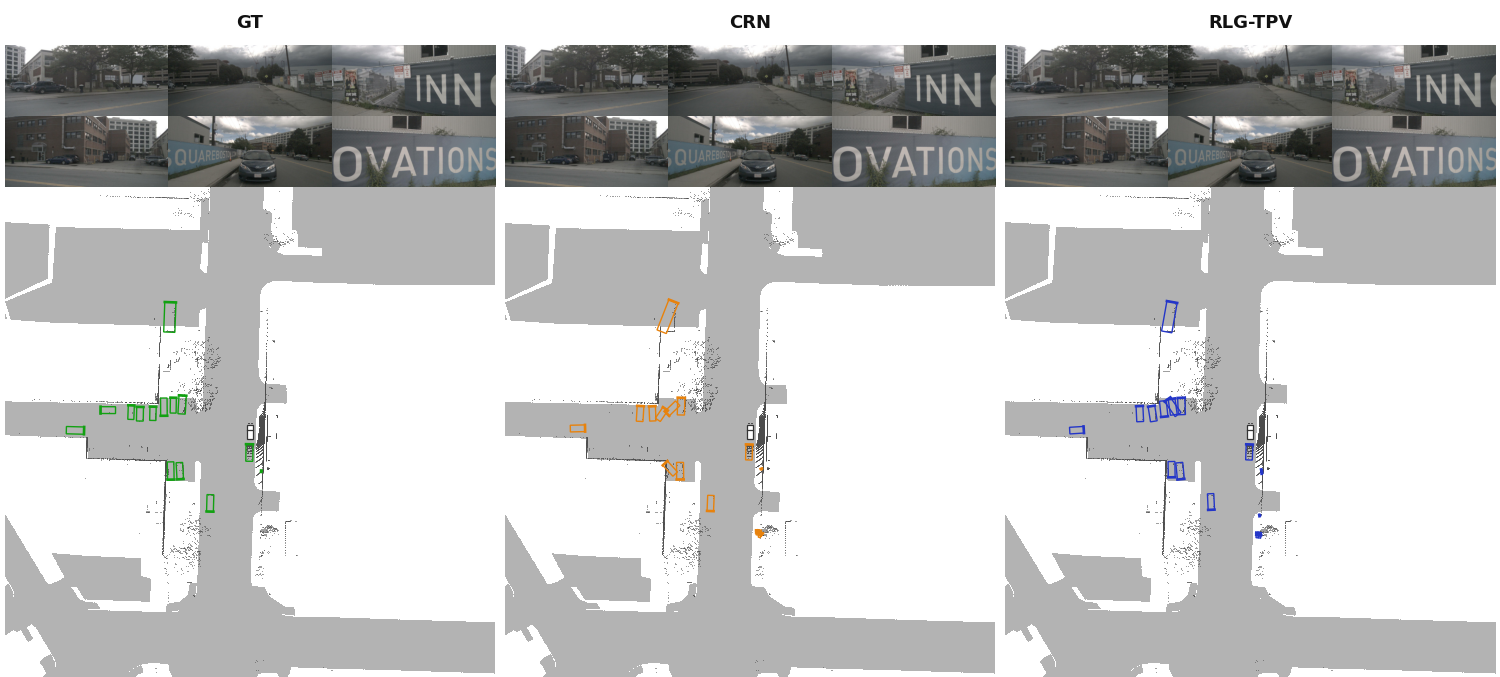}\\[0.45em]
\includegraphics[width=0.8\textwidth]{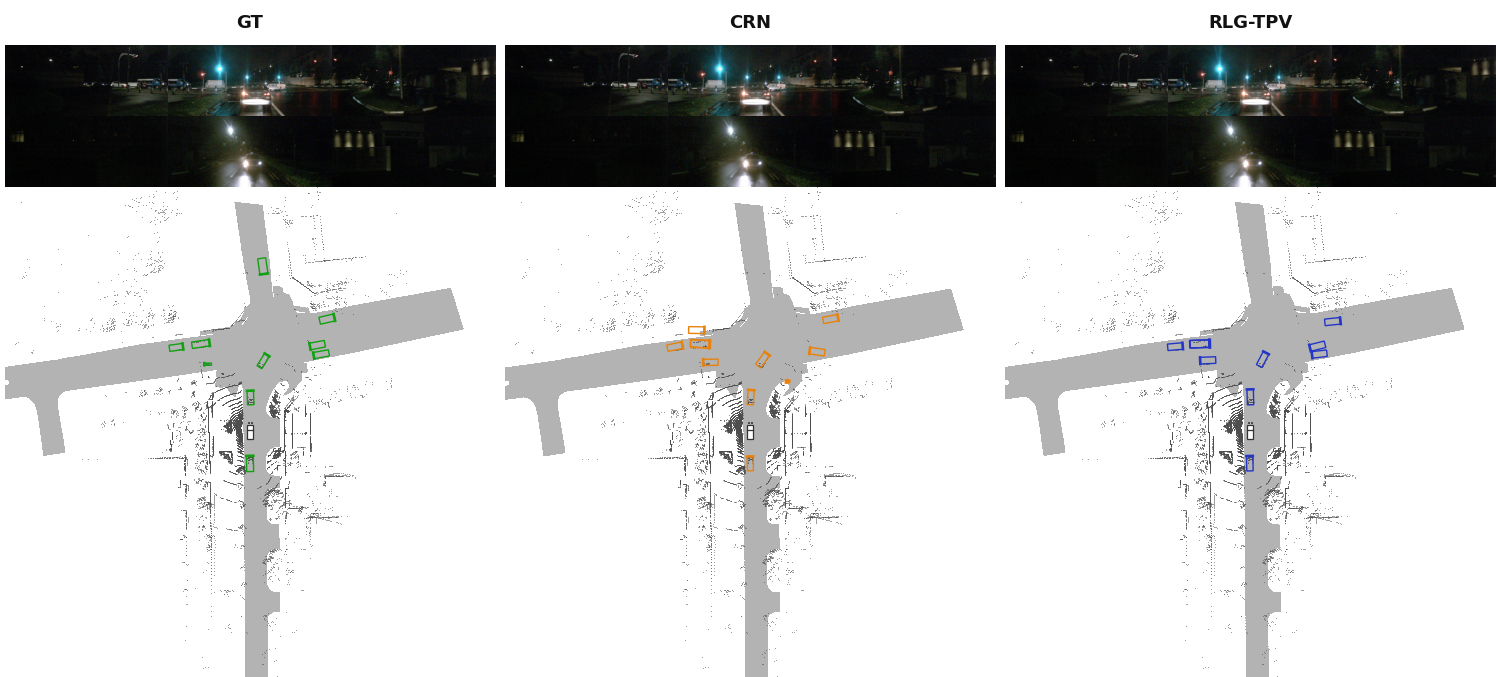}
\vspace{-0.15cm}
\caption{Qualitative comparison on nuScenes val. Columns: ground truth,
  CRN~\cite{crn}, RLG-TPV. Boxes are green (GT), orange (CRN) and blue (ours);
  map and LiDAR points are shown for visualisation only, as inference uses
  cameras and radar. \textbf{Top:} daytime crossing. \textbf{Bottom:} night.}
\label{fig:qual}
\end{figure*}

The magnitudes should nevertheless be interpreted conservatively. Several incremental changes \cref{tab:tpv} lie within the approximate $\pm0.005$ mAP variation observed across repeated runs, and one of three from-scratch reference runs did not converge normally. Moreover, the multi-plane row in \cref{tab:tpv} changes several coupled components simultaneously. The main experimental conclusions therefore rely on effects that are reproduced across complementary comparisons—most clearly the lift-development experiment and the removal of ray guidance—rather than on individual sub-percent increments.

\begin{table}[htbp]
\centering
\caption{Removal ablations on nuScenes val. Each row removes one component from the full model and reports the relative change after retraining.}
\vspace{-0.2cm}

\label{tab:abl}
\scriptsize
\setlength{\tabcolsep}{4pt}
\begin{tabular}{lcc}
\toprule
Configuration & $\Delta$ mAP & $\Delta$ NDS \\
\midrule
$-$ radar entirely (camera-only)
& $-35.5\%$ & $-30.2\%$ \\

$-$ temporal context (1 keyframe)
& $-11.5\%$ & $-18.8\%$ \\

$-$ ray guidance (unguided attention)
& $-7.3\%$ & $-7.2\%$ \\

$-$ sampling density ($K=4$)
& $-2.7\%$ & $-0.7\%$ \\

$-$ RCS-aware scatter
& $-2.1\%$ & $-4.7\%$ \\

$-$ radar-refined depth (RRD)
& $-0.1\%$ & $-1.4\%$ \\
\bottomrule
\end{tabular}
\end{table}
\vspace{-0.2cm}

In a complementary experiment we replaced the learned TPV height collapse with a substantially larger 3D-convolutional encoder, adding approximately $463$\,K parameters against approximately $2$\,K for the learned collapse; this reduces mAP by 0.0059 and NDS by 0.0031. The result suggests that retaining vertical structure is useful during representation
construction, but preserving explicit height resolution throughout the final detection head is not necessary in the configuration studied here. The multi-plane features can instead be aggregated before the conventional BEV detection head.

\subsection{Qualitative Comparison}
\label{sec:qual}
\vspace{-0.15cm}

\cref{fig:qual} compares the two models on a daytime and a nighttime scene. At the daytime crossing both models recover the row of parked vehicles along the side street, and the differences are concentrated in the consistency of the predicted orientations along that row and in isolated boxes placed in regions without a corresponding object. This is consistent with the aggregate error profile in \cref{tab:sota}, where the largest change relative to the CRN baseline occurs in orientation error. In the night scene the visible image evidence is limited to a small number of illuminated objects, and both models produce predictions in regions where the ground truth contains no annotated object; the qualitative comparison in this regime is therefore illustrative rather than conclusive. We use these examples to show the typical behavior of the two models and rely on the quantitative results for the comparison itself.

\section{Conclusion}
\label{sec:conclusion}
\vspace{-0.2cm}

We presented RLG-TPV, a multimodal tri-perspective representation for camera-radar 3D object detection that incorporates radar and training-time LiDAR geometry directly into the lifting process. The main component is a ray-guided deformable-attention lift in which LiDAR-supervised depth probabilities and radar occupancy weight individual samples along projected camera rays. RLG-TPV further uses measured radar Doppler to guide temporal alignment and employs LiDAR-derived occupancy to
constrain the side and front planes during training. On nuScenes val, RLG-TPV reaches 0.4981 mAP and 0.5959 NDS, with relative reductions of 31.9\% in orientation error and 30.7\% in velocity error compared with the published CRN baseline. Overall, the results indicate that the placement of geometric cues within representation construction is as important as the availability of the cues themselves. Rotation-aware height-plane targets
and extension to radar with elevation measurements remain directions for future work.


{
    \small
    \bibliographystyle{ieeenat_fullname}
    \bibliography{references}
}


\end{document}